\PassOptionsToPackage{table}{xcolor}
\documentclass[runningheads]{llncs}

\usepackage[T1]{fontenc}
\IfFileExists{newtxtext.sty}{%
  \usepackage{newtxtext}%
  \usepackage[varvw]{newtxmath}%
}{%
  \usepackage{mathptmx}%
  \usepackage{amsmath,amssymb}%
}
\usepackage{microtype}
\usepackage[table]{xcolor}
\usepackage{booktabs}
\usepackage{tabularx}
\usepackage{multirow}
\usepackage{algorithm}
\usepackage{algorithmic}
\usepackage{graphicx}
\usepackage{textcomp}
\usepackage{url}

\definecolor{LightPurple}{RGB}{248,240,255}
\definecolor{reda}{RGB}{192,0,0}

\usepackage{hyperref}
\hypersetup{
    colorlinks = true,
    urlcolor   = blue,
    linkcolor  = reda,
    citecolor  = blue
}

\begin{document}

\title{Multi-Scale Selective State-Space Memory for Temporal Knowledge Graph Extrapolation}

\titlerunning{Multi-Scale Selective State-Space Memory for TKG Extrapolation}

\author{Xitong Xu\inst{1} \and
Pingyang Huang\inst{1} \and
Yuqian Gao\inst{2} \and
Siyuan Li\inst{1}\thanks{Corresponding author.}}
\authorrunning{X. Xu et al.}
\institute{Dalian University of Technology, Dalian, China \and
Sun Yat-sen University, Guangzhou, China}

\maketitle

\begin{abstract}
Temporal knowledge graph (TKG) extrapolation forecasts future facts from strictly causal histories, where predictive cues span conflicting timescales: acute recent triggers, durable long-range regimes, and repeated subject--relation habits. Existing methods typically flatten history into a single temporal resolution and aggregate or copy past events passively, diluting immediate shifts with stale noise, truncating long-range regularities, or degrading sharply on novel entities. We propose \textbf{MS3M} (\textbf{M}ulti-\textbf{S}cale \textbf{S}elective \textbf{S}tate-\textbf{S}pace \textbf{M}emory), which reframes TKG extrapolation as multi-resolution selective memory consolidation paired with recurrence-calibrated open-space forecasting. MS3M first constructs a causal multi-scale pyramid via parameter-free, mask-aware coarsening to separate event-level triggers from coarse behavioral trajectories. Within each scale, an input-dependent diagonal selective state-space recurrence governs memory retention, writing, and readout exposure in strictly linear $O(dL)$ time, avoiding nonlinear RNN hidden-state coupling. Finally, query-conditioned attention retrieves multi-scale memory for full-entity generation, softly calibrated by a frequency- and recency-aware recurrence branch via a supervised adaptive gate. Across six benchmarks, MS3M leads Hits@1 on all six datasets and MRR on five (up to $+6.53$ Hits@1 on WIKI, $+5.67$ MRR on YAGO), with up to $+32.1\%$ relative MRR gains on novel entities.

\end{abstract}

\section{Introduction}

Temporal knowledge graph (TKG) extrapolation predicts future facts $(s,r,?,t)$ strictly from relational streams observed prior to timestamp $t$~\cite{cai2022temporal,hogan2021knowledge,zhang2024temporal}. As a core capability in temporal graph management and data mining, it underpins geopolitical risk forecasting, financial surveillance, dynamic recommendation, and query analysis~\cite{liu2026joint,liao2024gentkg,wang2024kgquestion,wang2025knowledge,wang2026beyond}. Unlike static link prediction and semantic-aware knowledge graph completion~\cite{li2026multi,li2025context,li2025flow,liu2025semantic}, future TKG facts emerge from non-stationary entity histories governed simultaneously by acute short-term triggers, durable long-range regularities, and repeated interaction habits~\cite{zhu2021learning,zhang2024temporal,li2026evolving,10.1145/3774904.3792581}.

Prior TKG extrapolation methods have progressed through recurrent encoders~\cite{jin2020recurrent,zhu2021learning}, temporal graph neural networks (TGNNs)~\cite{li2021temporal,li2022tirgn,liu2023retia}, temporal path and rule reasoners~\cite{han2020xerte,liu2022tlogic,huang2024confidence}, and contrastive or generative models~\cite{xu2023temporal,cai2024predicting,chen2025cogntke}. Despite steady aggregate gains, prevailing models still compress an entity's causal history into one temporal granularity, a fixed lookback window, or a rigid history-copying branch. This structural design creates a tightly coupled \emph{trilemma}:

\begin{figure}[t]
    \centering
    \includegraphics[width=0.96\linewidth]{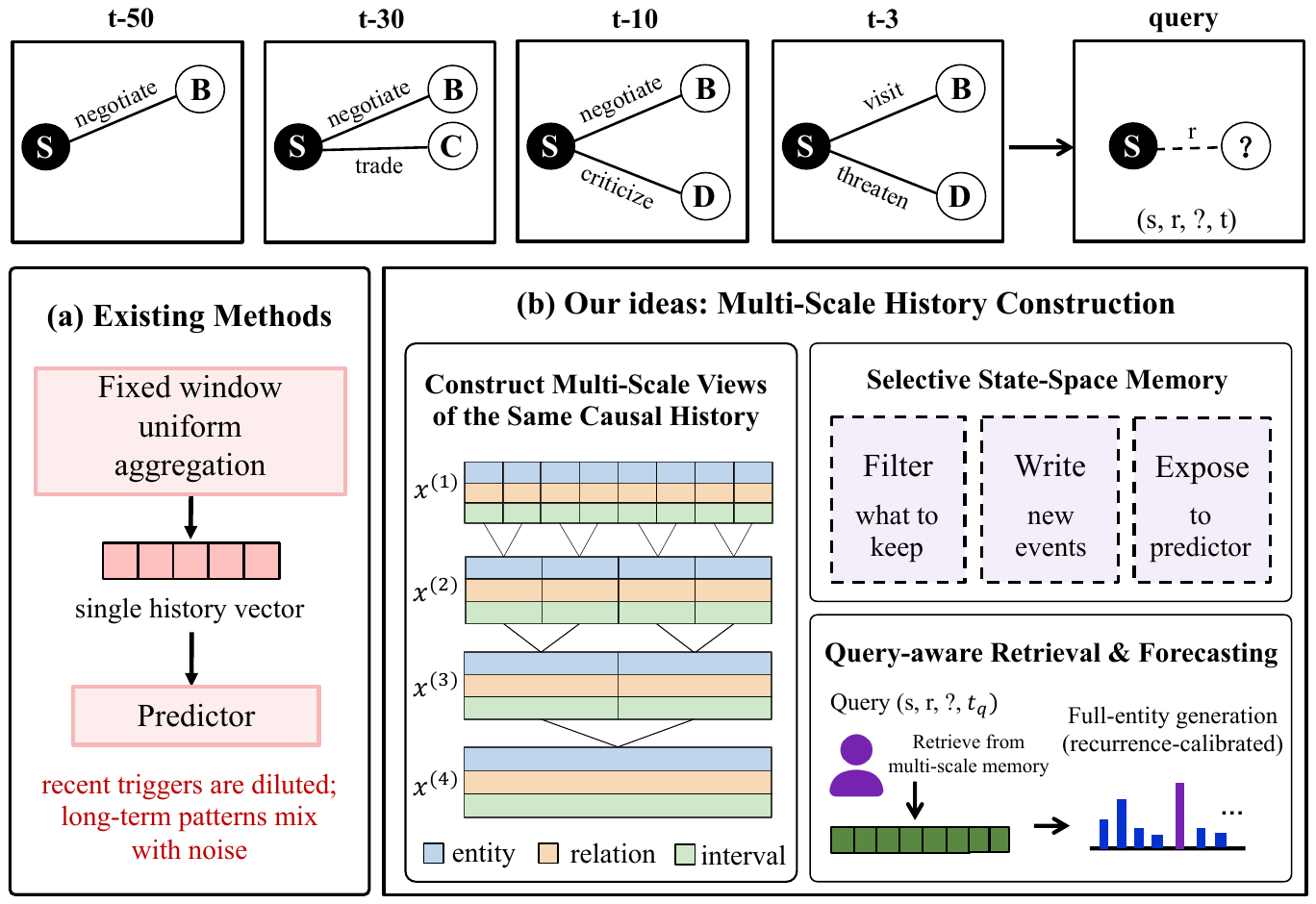}
    \caption{Conceptual contrast between single-scale aggregation and MS3M. Flattening causal history into one resolution dilutes acute triggers and entangles long-range regularities with noise. MS3M organizes the same causal stream into multi-scale selective state-space memories and retrieves query-relevant evidence for both recurrent and novel-entity forecasting.}
    \label{fig:motivation}
\end{figure}

\noindent\textbf{1. Single-Scale Resolution Mismatch.} Predictive evidence in TKGs spans inherently heterogeneous timescales~\cite{zhang2024survey}. As illustrated in Figure~\ref{fig:motivation}, an acute diplomatic clash immediately elevates fine-grained relations such as \textit{criticize} or \textit{threaten}, whereas durable alliances sustain coarse-grained regularities such as \textit{trade with} or \textit{negotiate with} across quiet intervals. Real-world queries routinely depend on both: forecasting an upcoming negotiation hinges on a recent escalation alongside historically stable partners. Collapsing history into a single resolution forces an unavoidable trade-off---wide windows dilute acute triggers with stale interactions, whereas narrow windows sever long-range behavioral regimes and recurring relational context.

\noindent\textbf{2. Passive Compression vs.\ Selective Memory Consolidation.} Multi-resolution views alone cannot suppress temporal noise if events are aggregated passively. At every scale, historical facts vary sharply in causal utility: some signal regime shifts, others act as transient triggers, and many are irrelevant to the target query. Uniform pooling, fixed-window aggregation, and standard message passing lack explicit input-dependent gating over \emph{which historical facts should be retained across time, written into latent memory, or exposed at query readout}.

\noindent\textbf{3. The Recurrence--Novelty Dilemma.} Because many TKG facts repeat under identical subject--relation contexts $(s,r)$, copy-centric predictors~\cite{zhu2021learning,xu2023temporal} exploit repetition effectively yet degrade sharply on novel (non-recurrent) tails---which constitute $38\%$--$43\%$ of test queries on standard benchmarks. Conversely, pure open-space decoders preserve novel-entity generalization but underutilize strong historical recurrence priors.

To resolve this coupled trilemma within a single scalable architecture, we propose \textbf{MS3M} (\textbf{M}ulti-\textbf{S}cale \textbf{S}elective \textbf{S}tate-\textbf{S}pace \textbf{M}emory). Rather than treating entity history as a flat sequence to be pooled or copied, MS3M reframes TKG extrapolation as \emph{multi-resolution selective memory consolidation} paired with \emph{recurrence-calibrated open-space forecasting}. Specifically, MS3M integrates three cohesive stages: (i) \emph{Causal Multi-Scale History Encoding} constructs a $K$-level temporal pyramid over each subject's strictly causal stream via parameter-free, mask-aware coarsening, separating acute event triggers from coarse behavioral trajectories; (ii) \emph{Diagonal Selective State-Space Memory} evolves each scale through input-dependent retention, writing, and readout exposure gates---instantiating a discretized first-order diagonal selective state-space model~\cite{gu2022efficiently,gu2024mamba} that eliminates GRU/LSTM hidden-state coupling and guarantees strictly linear $O(dL)$ memory-formation complexity---followed by cross-scale fusion and relation-conditioned readout; and (iii) \emph{Recurrence-Calibrated Open-Space Forecasting} retrieves query-relevant context via masked attention to score the full entity space $\mathcal{E}$, softly calibrated by a frequency- and recency-aware recurrence branch through an explicitly supervised adaptive gate that balances repetition against novelty.

Evaluated across six benchmarks---four dense event-centric streams (ICEWS14, ICEWS18, ICEWS05-15, GDELT) and two long-horizon graphs (WIKI, YAGO)---MS3M achieves the highest Hits@1 on all six datasets and the highest MRR on five, while delivering decisive gains on novel future entities. Our contributions are fourfold:
\begin{itemize}\setlength{\itemsep}{2pt}\setlength{\parskip}{0pt}
    \item \textbf{Multi-Scale Selective Memory Formulation:} We characterize resolution mismatch, passive aggregation, and the recurrence--novelty dilemma as coupled bottlenecks in TKG extrapolation, and reformulate temporal forecasting as unified multi-resolution selective memory consolidation.
    \item \textbf{Linear-Time Diagonal Selective SSM Architecture:} We propose \textbf{MS3M}, combining parameter-free, mask-aware causal coarsening with input-dependent diagonal linear recurrences that decouple latent consolidation from query readout in strictly linear $O(dL)$ memory-formation time (Section~\ref{sec:methodology}).
    \item \textbf{Recurrence-Calibrated Open-Space Forecasting:} We design a query-aware retrieval and adaptive calibration module that softly injects frequency- and recency-calibrated recurrence priors into open-space generation without suppressing non-recurrent tails (Section~\ref{sec:forecasting}).
    \item \textbf{Comprehensive Multi-Regime Validation:} Across six benchmarks, MS3M sets new top-rank precision records (up to $+6.53$ Hits@1 on WIKI and $+5.67$ MRR on YAGO), passes multi-seed significance tests ($p < 0.01$), yields up to $+32.1\%$ relative MRR gains on novel entities, and validates every core design choice against strong alternatives (Section~\ref{sec:exp}).
\end{itemize}

\section{Related Work}
\label{sec:related}

\subsection{Temporal Knowledge Graph Extrapolation.} Early TKG models extend static embeddings with time-aware hyperplanes or temporal projections~\cite{trivedi2017know,dasgupta2018hyte,garcia2018learning}. Extrapolation methods subsequently evolved along three primary paradigms: recurrent sequence encoders tracking entity trajectories~\cite{jin2020recurrent,zhu2021learning}, temporal GNNs propagating structural dependencies across discrete snapshots~\cite{li2021temporal,li2022complex,li2022tirgn,park2022evokg}, and temporal path or rule reasoners~\cite{han2020xerte,sun2021timetraveler,liu2022tlogic}. Recent advances enrich forecasting with relation-aware twin aggregation (RETIA~\cite{liu2023retia}), historical contrastive learning (CENET~\cite{xu2023temporal}), conditional diffusion (DiffuTKG~\cite{cai2024predicting}), time-varying rule confidence (TempValid~\cite{huang2024confidence}), cognitive global--local reasoning (CognTKE~\cite{chen2025cogntke}), and diachronic language encoding (DSEP~\cite{deng2025diachronic}). Nevertheless, most methods still compress history at a single granularity or rely on costly multi-hop snapshot propagation, struggling to preserve acute triggers and long-range regimes simultaneously at scale. MS3M instead constructs and consolidates a compact multi-resolution causal hierarchy in strictly linear $O(dL)$ time without multi-hop snapshot expansion.

\subsection{Multi-Granularity History, Recurrence, and Selective State Spaces.} Several studies exploit recurrence priors or multi-granularity signals to capture heterogeneous temporal cues. CyGNet~\cite{zhu2021learning} pioneers sequential copy-generation for repeated facts, while CENET~\cite{xu2023temporal} contrasts historical against non-historical entities. Local--global and multi-granularity frameworks---including TiRGN~\cite{li2022tirgn}, RLGNet~\cite{rlgnet2024}, MGESL~\cite{mgesl2024}, and CognTKE~\cite{chen2025cogntke}---combine recent interactions with global history or entity similarity signals. MS3M differs fundamentally along three axes: (1) rather than heuristic local/global branches or static similarity tables, MS3M builds a unified $K$-level causal pyramid via mask-aware coarsening over the same event stream; (2) rather than passive pooling or nonlinearly coupled GRU/LSTM cells~\cite{hochreiter1997long}, MS3M instantiates \emph{diagonal selective state-space recurrences}~\cite{gu2022efficiently,gu2024mamba} that independently regulate retention, writing, and readout exposure; and (3) rather than hard copying that penalizes non-recurrent tails, MS3M calibrates recurrent logits by frequency and recency lag and blends them softly with open-space generation via a supervised adaptive gate.

\section{Methodology}
\label{sec:methodology}

\subsection{Problem Formulation and Overview}
\label{sec:method_overview}

A temporal knowledge graph $\mathcal{G} \subseteq \mathcal{E} \times \mathcal{R} \times \mathcal{E} \times \mathcal{T}$ consists of timestamped quadruples $(s,r,o,t)$ over entity set $\mathcal{E}$, relation set $\mathcal{R}$, and timestamp set $\mathcal{T}$. Given a future query $(s,r,?,t)$, TKG extrapolation ranks all candidate tails $o \in \mathcal{E}$ strictly from causal history $\mathcal{G}_{<t} = \{(s',r',o',t') \in \mathcal{G} \mid t' < t\}$. Head prediction $(?,r,o,t)$ is handled symmetrically via inverse relations as $(o,r^{-1},?,t)$.

Algorithm~\ref{alg:ms3m} summarizes MS3M's end-to-end inference across three stages: (1) \emph{Causal Multi-Scale History Encoding} (Section~\ref{sec:history_encoding}) builds a $K$-level temporal pyramid $\{\mathbf{X}^{(k)}\}_{k=1}^K$ from the subject's causal stream $\mathcal{H}_{s,t}$; (2) \emph{Selective State-Space Memory Formation} (Section~\ref{sec:memory_formation}) runs input-dependent diagonal linear recurrences at each scale, fuses aligned readouts into $\mathbf{Z}$, and applies relation-conditioned gating to obtain $\bar{\mathbf{Z}}$; and (3) \emph{Recurrence-Calibrated Forecasting} (Section~\ref{sec:forecasting}) retrieves query-relevant context $\mathbf{c}$ to score the full entity space $P_{\mathrm{gen}}$ and softly calibrates it with a frequency- and recency-aware recurrence distribution $P_{\mathrm{rec}}$ via an adaptive gate $\lambda$.

\subsection{Causal Multi-Scale History Encoding}
\label{sec:history_encoding}

For query $(s,r,?,t)$, MS3M extracts the subject's strictly causal event stream:
\begin{equation}
\mathcal{H}_{s,t} = \{(r_i,o_i,t_i)\mid (s,r_i,o_i,t_i)\in\mathcal{G},\ t_i<t\},
\end{equation}
retaining the $L$ most recent events chronologically with a binary validity mask $\mathbf{m}^{(1)} \in \{0,1\}^L$ (abbreviated $\mathbf{m}$) to pad shorter histories without leaking future timestamps.

Each event $(r_i, o_i, t_i)$ encodes both relational context and query-relative temporal proximity. We project its relative lag $\Delta_i = t - t_i$ through a learnable temporal encoder $\mathbf{d}_i = \phi(\Delta_i)$, and modulate neighbor embedding $\mathbf{e}_{o_i}$ by its interacting relation $\mathbf{e}_{r_i}$ via relation-conditioned residual layer normalization:
\begin{equation}
\tilde{\mathbf{e}}_{o_i} = \mathrm{LN}\!\left(\mathbf{e}_{o_i} + \alpha\mathbf{W}_{r}\mathbf{e}_{r_i}\right),
\end{equation}
where $\mathbf{W}_{r}$ is a learnable projection and $\alpha$ is a learnable scalar. Concatenating these components forms event token $\mathbf{x}_i = [\tilde{\mathbf{e}}_{o_i};\,\mathbf{e}_{r_i};\,\mathbf{d}_i]$, yielding the finest-scale sequence $\mathbf{X}^{(1)} = [\mathbf{x}_1,\ldots,\mathbf{x}_L]$.

To disentangle acute triggers from coarse behavioral trajectories without adding per-scale convolution parameters, MS3M recursively derives coarser views $\{\mathbf{X}^{(k)}\}_{k=1}^{K}$ via mask-aware temporal coarsening with downsampling factor $\rho$:
\begin{equation}
\mathbf{x}^{(k+1)}_j = \frac{\sum_{i\in\mathcal{B}_j} m^{(k)}_i\mathbf{x}^{(k)}_i}{\sum_{i\in\mathcal{B}_j} m^{(k)}_i + \epsilon}, \qquad
m^{(k+1)}_j = \mathbb{I}\!\left[\sum_{i\in\mathcal{B}_j} m^{(k)}_i > 0\right],
\end{equation}
where $\mathcal{B}_j$ indexes the contiguous $\rho$-step window at scale $k$ aggregated into position $j$ at scale $k+1$, and $\epsilon > 0$ prevents division by zero. Fine scales preserve event-level triggers while coarse scales summarize persistent interaction regimes over the exact same causal history (empirically validated against dilated, convolutional, and time-binned coarsening alternatives in Section~\ref{sec:ablation}).

\subsection{Selective State-Space Memory Formation}
\label{sec:memory_formation}

While $\{\mathbf{X}^{(k)}\}_{k=1}^K$ exposes multi-resolution structure, passive aggregation within each scale still entangles durable regularities with transient noise. MS3M therefore maintains a selective state-space memory stream per scale. Given token $\mathbf{x}^{(k)}_i$ at scale $k$, we project it into memory input $\mathbf{u}^{(k)}_i = \mathbf{W}_{u}\mathbf{x}^{(k)}_i$ and derive a local control representation $\mathbf{p}^{(k)}_i = \psi(\mathbf{u}^{(k)}_i)$ (implemented via a lightweight depthwise causal convolution with channel expansion), producing three input-dependent gates:
\begin{equation}
\mathbf{a}^{(k)}_i = \sigma(\mathbf{b}_a + \mathbf{W}_a\mathbf{p}^{(k)}_i), \quad
\mathbf{w}^{(k)}_i = \sigma(\mathbf{W}_w\mathbf{p}^{(k)}_i), \quad
\mathbf{g}^{(k)}_i = \sigma(\mathbf{W}_g\mathbf{p}^{(k)}_i),
\end{equation}
which govern \emph{retention} ($\mathbf{a}^{(k)}_i$), \emph{writing} ($\mathbf{w}^{(k)}_i$), and \emph{readout exposure} ($\mathbf{g}^{(k)}_i$), respectively. Latent memory $\mathbf{h}^{(k)}_i$ evolves along scale $k$ while exposed readout $\mathbf{z}^{(k)}_i$ is gated independently with a learnable diagonal skip vector $\mathbf{s}$:
\begin{align}
\mathbf{h}^{(k)}_i &= \mathbf{a}^{(k)}_i\odot\mathbf{h}^{(k)}_{i-1} + \mathbf{w}^{(k)}_i\odot\tanh(\mathbf{u}^{(k)}_i), \label{eq:ssm_state} \\
\mathbf{z}^{(k)}_i &= \mathbf{g}^{(k)}_i\odot\mathbf{h}^{(k)}_i + \mathbf{s}\odot\mathbf{u}^{(k)}_i. \label{eq:ssm_out}
\end{align}

\noindent\textbf{Connection to Selective State-Space Models.} A continuous-time linear time-varying SSM~\cite{gu2022efficiently,gu2024mamba} evolves latent state via $\dot{\mathbf{h}}(t) = \mathbf{A}(t)\mathbf{h}(t) + \mathbf{B}(t)\mathbf{u}(t)$ and $\mathbf{z}(t) = \mathbf{C}(t)\mathbf{h}(t) + \mathbf{D}\mathbf{u}(t)$. Parameterizing $\mathbf{A}(t)$ as negative diagonal and discretizing with input-dependent step sizes yields the diagonal selective recurrence $\mathbf{h}_i = \bar{\mathbf{A}}_i \mathbf{h}_{i-1} + \bar{\mathbf{B}}_i \tilde{\mathbf{u}}_i$, $\mathbf{z}_i = \mathbf{C}_i \mathbf{h}_i + \mathbf{D}\mathbf{u}_i$. Equations~\eqref{eq:ssm_state}--\eqref{eq:ssm_out} directly instantiate this first-order diagonal selective SSM: $\bar{\mathbf{A}}_i = \mathrm{diag}(\mathbf{a}^{(k)}_i) \in (0,1)^d$ enforces bounded-input bounded-output (BIBO) stability via contractive decay, $\bar{\mathbf{B}}_i = \mathrm{diag}(\mathbf{w}^{(k)}_i)$ modulates bounded inputs $\tanh(\mathbf{u}^{(k)}_i)$, $\mathbf{C}_i = \mathrm{diag}(\mathbf{g}^{(k)}_i)$ controls selective exposure, and $\mathbf{D} = \mathrm{diag}(\mathbf{s})$ preserves acute cues via direct feedthrough. Unlike classical GRU/LSTM cells~\cite{hochreiter1997long}, whose gates depend nonlinearly on $\mathbf{h}_{i-1}$ via dense projections $\mathbf{U}\mathbf{h}_{i-1}$ and break associativity, $(\mathbf{a}^{(k)}_i, \mathbf{w}^{(k)}_i, \mathbf{g}^{(k)}_i)$ depend \emph{exclusively on the input stream}. Equation~\eqref{eq:ssm_state} is thus an input-dependent diagonal linear recurrence supporting $O(\log L)$-depth parallel associative scans, stabilized gradients, and decoupled latent consolidation versus readout.

\noindent\textbf{Cross-Scale Fusion and Relation-Conditioned Readout.} MS3M aligns coarser readouts back to the finest grid and fuses them into $\mathbf{Z} = [\mathbf{z}_1,\ldots,\mathbf{z}_L]$ via $\mathbf{z}_i = \sum_{k=1}^{K} \omega_k\hat{\mathbf{z}}^{(k)}_i$, where $\hat{\mathbf{z}}^{(k)}_i$ is the aligned scale-$k$ readout. We use uniform weights ($\omega_k = 1/K$) by default to avoid redundant parameterization, as per-scale exposure gates and downstream query attention already perform adaptive routing (verified in Section~\ref{sec:ablation}). Because query relations rely asymmetrically on consolidated memory versus raw event triggers (e.g., enduring treaties vs.\ abrupt escalations), MS3M balances both via a relation-conditioned gate $\boldsymbol{\eta}_{r} = \sigma(\mathbf{W}_{\eta}\mathbf{e}_{r})$:
\begin{equation}
\bar{\mathbf{z}}_i = \boldsymbol{\eta}_{r}\odot\tanh(\mathbf{z}_i) + (1-\boldsymbol{\eta}_{r})\odot\tanh(\mathbf{x}'_i),
\end{equation}
where $\mathbf{x}'_i = \mathbf{W}_{x}\mathbf{x}^{(1)}_i$ is the projected fine-grained token, yielding $\bar{\mathbf{Z}} = [\bar{\mathbf{z}}_1,\ldots,\bar{\mathbf{z}}_L]$.

\subsection{Recurrence-Calibrated Memory Retrieval and Forecasting}
\label{sec:forecasting}

Given query vector $\mathbf{q} = \mathbf{W}_{q}[\mathbf{e}_{s};\,\mathbf{e}_{r};\,\phi(t)]$, MS3M retrieves query-relevant evidence from $\bar{\mathbf{Z}}$ via masked attention:
\begin{equation}
\pi_i = \frac{\exp(\mathbf{w}_{\pi}^{\top}[\bar{\mathbf{z}}_i;\,\mathbf{q}])}{\sum_{j:m_j=1}\exp(\mathbf{w}_{\pi}^{\top}[\bar{\mathbf{z}}_j;\,\mathbf{q}])}, \qquad
\mathbf{c} = \sum_{i:m_i=1}\pi_i\bar{\mathbf{z}}_i.
\end{equation}

\noindent\textbf{Open-Space Full-Entity Generation.} To preserve generalization on non-recurrent facts, MS3M scores every candidate $o \in \mathcal{E}$ across the full entity vocabulary:
\begin{align}
S_{\mathrm{gen}}(o\mid s,r,t) &= (\mathbf{c}\odot\mathbf{e}_{r})^{\top}(\mathbf{e}_{o}+\mathbf{e}^{\mathrm{obj}}_{o}), \\
P_{\mathrm{gen}}(o\mid s,r,t) &= \frac{\exp(S_{\mathrm{gen}}(o\mid s,r,t))}{\sum_{o'\in\mathcal{E}}\exp(S_{\mathrm{gen}}(o'\mid s,r,t))},
\end{align}
where $\mathbf{e}^{\mathrm{obj}}_{o}$ is a learnable object-role embedding.

\noindent\textbf{Frequency- and Recency-Calibrated Recurrence.} Let $\mathcal{C}_{s,r,t} = \{o \mid (s,r,o,t') \in \mathcal{G},\ t' < t\}$ denote entities previously observed under $(s,r)$. Rather than blindly copying $\mathcal{C}_{s,r,t}$, MS3M calibrates each recurrent candidate's generation logit by its historical frequency $f_{s,r,o}$ and recency lag $\Delta_{s,r,o}$ (time since latest occurrence):
\begin{equation}
S_{\mathrm{rec}}(o\mid s,r,t) = S_{\mathrm{gen}}(o\mid s,r,t) + \beta_f\log(1+f_{s,r,o}) - \beta_{\Delta}\log(1+\Delta_{s,r,o}),
\end{equation}
with learnable weights $\beta_f, \beta_{\Delta}$. Normalizing over $\mathcal{C}_{s,r,t}$ gives the recurrence distribution:
\begin{equation}
P_{\mathrm{rec}}(o\mid s,r,t) = \frac{\mathbb{I}[o\in\mathcal{C}_{s,r,t}]\exp(S_{\mathrm{rec}}(o\mid s,r,t))}{\sum_{o'\in\mathcal{C}_{s,r,t}}\exp(S_{\mathrm{rec}}(o'\mid s,r,t))},
\end{equation}
falling back to $P_{\mathrm{gen}}(o\mid s,r,t)$ when $\mathcal{C}_{s,r,t} = \emptyset$.

\noindent\textbf{Adaptive Fusion.} MS3M blends both branches through a query-conditioned recurrence gate $\lambda = \sigma(\mathbf{w}_{\lambda}^{\top}[\mathbf{q};\,\mathbf{c};\,\mathbf{e}_{r}])$:
\begin{equation}
P(o\mid s,r,t) = \lambda P_{\mathrm{rec}}(o\mid s,r,t) + (1-\lambda)P_{\mathrm{gen}}(o\mid s,r,t).
\end{equation}
This soft calibration concentrates mass onto causally supported recurrent tails while preserving open-space probability mass for novel entities.

\begin{algorithm}[t]
\caption{MS3M Extrapolation Procedure for a Temporal Query}
\label{alg:ms3m}
\small
\begin{algorithmic}[1]
\REQUIRE Causal history $\mathcal{G}_{<t}$, query $(s,r,?,t)$, history length $L$, scales $K$
\ENSURE Predictive entity distribution $P(o\mid s,r,t)$
\STATE Extract the $L$ most recent causal events $\mathcal{H}_{s,t}$ ($t_i < t$) with validity mask $\mathbf{m}$, and encode relation-modulated, time-aware tokens $\mathbf{X}^{(1)}$.
\FOR{$k=1$ to $K-1$}
    \STATE Derive coarse view $\mathbf{X}^{(k+1)}$ and mask $\mathbf{m}^{(k+1)}$ via mask-aware coarsening.
\ENDFOR
\FOR{each scale $k=1,\ldots,K$}
    \STATE Run diagonal selective SSM updates ($\mathbf{a}_i^{(k)}, \mathbf{w}_i^{(k)}, \mathbf{g}_i^{(k)}$) to obtain $\{\mathbf{z}^{(k)}_i\}$ and align to the finest grid $\{\hat{\mathbf{z}}^{(k)}_i\}$.
\ENDFOR
\STATE Fuse aligned scales into $\mathbf{Z}$ and apply relation gate $\boldsymbol{\eta}_r$ to obtain $\bar{\mathbf{Z}}$.
\STATE Retrieve query context $\mathbf{c}$ via masked attention and score $P_{\mathrm{gen}}$ over $\mathcal{E}$.
\STATE Build $\mathcal{C}_{s,r,t}$, compute calibrated $P_{\mathrm{rec}}$, and output $P = \lambda P_{\mathrm{rec}} + (1-\lambda)P_{\mathrm{gen}}$.
\end{algorithmic}
\end{algorithm}

\subsection{Optimization and Complexity Analysis}
\label{sec:optimization}

MS3M is trained end-to-end by minimizing tail negative log-likelihood regularized by explicit binary cross-entropy supervision on the recurrence gate $\lambda$:
\begin{equation}
\mathcal{L} = \underbrace{-\log P(o^{*}\mid s,r,t)}_{\mathcal{L}_{\mathrm{pred}}} + \lambda_{\mathrm{rec}}\underbrace{\mathrm{BCE}(\lambda,\, y_{\mathrm{rec}})}_{\mathcal{L}_{\mathrm{rec}}}, \;\; y_{\mathrm{rec}} = \mathbb{I}[o^{*}\in\mathcal{C}_{s,r,t}],
\end{equation}
where $\lambda_{\mathrm{rec}}$ balances gate calibration against ranking accuracy.

\noindent\textbf{Strictly Linear Memory-Formation Complexity.} For history length $L$, hidden dimension $d$, $K$ scales, and coarsening factor $\rho > 1$, MS3M evaluates $\sum_{k=0}^{K-1} L/\rho^k$ diagonal state-space steps prior to entity scoring. Since $\sum_{k=0}^{K-1} \rho^{-k} < \frac{\rho}{\rho-1}$, total memory-formation complexity is strictly linear:
\begin{equation}
O\!\left(d\sum_{k=0}^{K-1}\frac{L}{\rho^k}\right) = O(dL).
\end{equation}
Under $\rho = 2$, multi-scale consolidation requires strictly less than $2\times$ the compute of a single-scale pass, avoiding both $O(dL^2)$ self-attention scaling and heavy multi-hop snapshot propagation---enabling seamless scaling to million-edge temporal graph streams such as GDELT under a single-GPU budget.

\section{Experiments}
\label{sec:exp}

We evaluate MS3M across six benchmarks to answer four research questions:
\begin{itemize}\setlength{\itemsep}{1pt}\setlength{\parskip}{0pt}
    \item \textbf{RQ1 (Overall Effectiveness):} How does MS3M perform against state-of-the-art TKG extrapolation baselines across event-centric and long-horizon regimes?
    \item \textbf{RQ2 (Ablation \& Design Verification):} What does each component contribute, and how do our coarsening, fusion, and diagonal SSM choices compare against competitive alternatives?
    \item \textbf{RQ3 (Hyperparameter Sensitivity):} How sensitive is MS3M to history length $L$, temporal scales $K$, and memory state dimension?
    \item \textbf{RQ4 (Heterogeneous Evidence Diagnosis):} Does MS3M simultaneously excel on short-term triggers, long-term recurrences, and novel non-recurrent entities?
\end{itemize}

\subsection{Experimental Setup}
\label{sec:experimental_setup}

\noindent\textbf{Datasets.} We evaluate on two complementary benchmark suites (Table~\ref{tab:dataset_stats}). The \emph{event-centric suite} includes ICEWS14~\cite{garcia2018learning}, ICEWS18~\cite{jin2020recurrent}, ICEWS05-15~\cite{li2021temporal}, and GDELT~\cite{leetaru2013gdelt}, which exhibit dense temporal dynamics combining acute short-term triggers, long-range regularities, and repeated $(s,r)$ habits. The \emph{long-horizon suite} comprises WIKI~\cite{leblay2018deriving} and YAGO~\cite{mahdisoltani2014yago3}, featuring coarser temporal granularity and persistent facts. All datasets follow standard chronological splits under strict causal extrapolation ($t_i < t$).

\begin{table}[t]
\centering
\footnotesize
\setlength{\tabcolsep}{6.5pt}
\renewcommand{\arraystretch}{1.05}
\caption{Statistics of the six TKG extrapolation benchmarks.}
\label{tab:dataset_stats}
\begin{tabular}{l|rrrrr}
\toprule
\textbf{Dataset} & \textbf{\#Ent.} & \textbf{\#Rel.} & \textbf{Train} & \textbf{Valid} & \textbf{Test} \\
\midrule
ICEWS14    & 7,128  & 230 & 74,845    & 8,514   & 7,371 \\
ICEWS18    & 23,033 & 256 & 373,018   & 45,995  & 49,545 \\
ICEWS05-15 & 10,488 & 251 & 368,868   & 46,302  & 46,159 \\
GDELT      & 7,691  & 240 & 1,734,399 & 238,765 & 305,241 \\
WIKI       & 12,554 & 24  & 539,286   & 67,538  & 63,110 \\
YAGO       & 10,623 & 10  & 161,540   & 19,523  & 20,026 \\
\bottomrule
\end{tabular}
\end{table}

\noindent\textbf{Baselines.} Following standard suite-specific protocols, on ICEWS/GDELT we compare against 13 baselines spanning recurrent models (RE-NET~\cite{jin2020recurrent}, CyGNet~\cite{zhu2021learning}), path reasoners (xERTE~\cite{han2020xerte}, TITer~\cite{sun2021timetraveler}), temporal GNNs (RE-GCN~\cite{li2021temporal}, CEN~\cite{li2022complex}, TiRGN~\cite{li2022tirgn}, RETIA~\cite{liu2023retia}, CENET~\cite{xu2023temporal}), and recent state-of-the-art architectures (DiffuTKG~\cite{cai2024predicting}, TempValid~\cite{huang2024confidence}, CognTKE~\cite{chen2025cogntke}, DSEP~\cite{deng2025diachronic}). On WIKI/YAGO, we compare against TeMP~\cite{wu2020temp}, RE-NET~\cite{jin2020recurrent}, xERTE~\cite{han2020xerte}, RE-GCN~\cite{li2021temporal}, TANGO-TuckER/DistMult~\cite{han2021temporal}, CyGNet~\cite{zhu2021learning}, EvoKG~\cite{park2022evokg}, HIP~\cite{he2024hip}, CENET~\cite{xu2023temporal}, and TempValid~\cite{huang2024confidence}.

\noindent\textbf{Protocol and Implementation.} We report time-aware filtered MRR and Hits@$K$ ($K\in\{1,3,10\}$) for bidirectional entity prediction. Unless noted otherwise, MS3M uses a unified configuration across all benchmarks: entity/relation dimension $64$, time dimension $32$, history length $L=32$, $K=4$ scales ($\rho=2$), selective state dimension $64$, causal convolution width $4$, and channel expansion factor $2$. Models are trained with Adam for $20$ epochs (batch size $128$, learning rate $5\times 10^{-3}$ with $2$-epoch warmup and cosine decay, weight decay $10^{-4}$) on one NVIDIA RTX 4090 GPU (24\,GB), selecting checkpoints by validation MRR. We verify multi-seed stability across $3$ seeds ($\{42, 1024, 2025\}$) using one-tailed one-sample $t$-tests against the strongest reported baseline.

\begin{table}[t]
\centering
\footnotesize
\setlength{\tabcolsep}{3pt}
\renewcommand{\arraystretch}{1.1}
\caption{Overall performance on four event-centric benchmarks. Best results are in \textbf{bold}; second best are \underline{underlined}. $^{\dagger}$ denotes statistically significant gain over the strongest baseline ($p < 0.01$, one-tailed one-sample $t$-test across $3$ seeds; see Table~\ref{tab:significance}).}
\label{tab:main_result}
\resizebox{\textwidth}{!}{%
\begin{tabular}{@{}l|cccc|cccc|cccc|cccc@{}}
\toprule
\multirow{2}{*}{\textbf{Model}} 
& \multicolumn{4}{c}{\textbf{ICEWS14}} 
& \multicolumn{4}{c}{\textbf{ICEWS18}} 
& \multicolumn{4}{c}{\textbf{ICEWS05-15}} 
& \multicolumn{4}{c}{\textbf{GDELT}} \\
\cmidrule(lr){2-5}\cmidrule(lr){6-9}\cmidrule(lr){10-13}\cmidrule(l){14-17}
& \textbf{MRR} & \textbf{H@1} & \textbf{H@3} & \textbf{H@10}
& \textbf{MRR} & \textbf{H@1} & \textbf{H@3} & \textbf{H@10}
& \textbf{MRR} & \textbf{H@1} & \textbf{H@3} & \textbf{H@10}
& \textbf{MRR} & \textbf{H@1} & \textbf{H@3} & \textbf{H@10} \\
\midrule
RE-NET (2020)       & 0.369 & 0.268 & 0.395 & 0.548 & 0.288 & 0.191 & 0.324 & 0.475 & 0.433 & 0.334 & 0.478 & 0.631 & 0.196 & 0.124 & 0.210 & 0.340 \\
CyGNet (2020)       & 0.351 & 0.257 & 0.390 & 0.536 & 0.249 & 0.159 & 0.283 & 0.426 & 0.368 & 0.266 & 0.416 & 0.562 & 0.185 & 0.115 & 0.196 & 0.320 \\
xERTE (2021)        & 0.400 & 0.321 & 0.446 & 0.562 & 0.293 & 0.210 & 0.335 & 0.465 & 0.466 & 0.378 & 0.523 & 0.639 & 0.181 & 0.123 & 0.201 & 0.303 \\
TITer (2021)        & 0.409 & 0.323 & 0.455 & 0.571 & 0.300 & 0.221 & 0.335 & 0.448 & 0.477 & 0.380 & 0.529 & 0.658 & 0.155 & 0.110 & 0.156 & 0.243 \\
RE-GCN (2021)       & 0.404 & 0.307 & 0.450 & 0.592 & 0.306 & 0.210 & 0.343 & 0.488 & 0.480 & 0.373 & 0.539 & 0.683 & 0.196 & 0.124 & 0.209 & 0.337 \\
CEN (2022)          & 0.422 & 0.321 & 0.475 & 0.613 & 0.315 & 0.217 & 0.354 & 0.506 & 0.468 & 0.364 & 0.524 & 0.670 & 0.204 & 0.130 & 0.218 & 0.350 \\
TiRGN (2022)        & 0.440 & 0.338 & 0.490 & 0.638 & 0.337 & 0.232 & 0.380 & 0.542 & 0.500 & 0.393 & 0.561 & 0.707 & 0.217 & 0.136 & 0.233 & 0.376 \\
RETIA (2023)        & 0.428 & 0.323 & 0.478 & 0.628 & 0.324 & 0.222 & 0.365 & 0.529 & 0.473 & 0.366 & 0.529 & 0.678 & 0.201 & 0.128 & 0.215 & 0.345 \\
CENET (2023)        & 0.390 & 0.296 & 0.432 & 0.575 & 0.279 & 0.182 & 0.316 & 0.470 & 0.420 & 0.322 & 0.469 & 0.604 & 0.202 & 0.127 & 0.217 & 0.349 \\
DiffuTKG (2024)     & 0.432 & 0.338 & 0.476 & 0.615 & 0.315 & 0.220 & 0.353 & 0.503 & 0.456 & 0.361 & 0.505 & 0.639 & 0.211 & 0.139 & 0.224 & 0.352 \\
TempValid (2024)    & \underline{0.458} & \underline{0.355} & \underline{0.513} & \textbf{0.651} & 0.335 & 0.239 & 0.379 & 0.523 & 0.503 & 0.395 & 0.567 & 0.706 & 0.219 & 0.144 & 0.244 & 0.370 \\
CognTKE (2025)      & 0.450 & 0.353 & 0.509 & 0.636 & \underline{0.342} & \underline{0.240} & 0.381 & \textbf{0.553} & \textbf{0.519} & \underline{0.408} & \textbf{0.584} & \textbf{0.721} & \underline{0.241} & \underline{0.153} & \underline{0.258} & \textbf{0.430} \\
DSEP (2025)         & 0.449 & 0.344 & 0.503 & \underline{0.649} & 0.338 & 0.232 & \underline{0.382} & \underline{0.545} & 0.507 & 0.398 & 0.568 & \underline{0.713} & 0.220 & 0.139 & 0.237 & 0.381 \\
\midrule
\textbf{MS3M}       & \textbf{0.478}$^{\dagger}$ & \textbf{0.391}$^{\dagger}$ & \textbf{0.520} & 0.642 & \textbf{0.358}$^{\dagger}$ & \textbf{0.262}$^{\dagger}$ & \textbf{0.403}$^{\dagger}$ & 0.539 & \underline{0.517} & \textbf{0.421}$^{\dagger}$ & \underline{0.570} & 0.698 & \textbf{0.243}$^{\dagger}$ & \textbf{0.159}$^{\dagger}$ & \textbf{0.266}$^{\dagger}$ & \underline{0.407} \\
\bottomrule
\end{tabular}%
}
\end{table}

\begin{table}[t]
\centering
\footnotesize
\setlength{\tabcolsep}{7.5pt}
\renewcommand{\arraystretch}{1.0}
\caption{Overall results (\%) on WIKI and YAGO (best in \textbf{bold}; second best \underline{underlined}).}
\label{tab:wiki_yago_results}
\begin{tabular}{l|ccc|ccc}
\toprule
\multirow{2}{*}{\textbf{Method}} 
& \multicolumn{3}{c|}{\textbf{WIKI}} 
& \multicolumn{3}{c}{\textbf{YAGO}} \\
\cmidrule(lr){2-4} \cmidrule(lr){5-7}
& \textbf{MRR} & \textbf{Hits@1} & \textbf{Hits@3}
& \textbf{MRR} & \textbf{Hits@1} & \textbf{Hits@3} \\
\midrule
TeMP          & 49.61 & 46.96 & 50.24 & 62.25 & 55.39 & 64.63 \\
RE-NET        & 51.97 & 48.01 & 52.07 & 65.16 & 63.29 & 65.63 \\
xERTE         & $>$1\,day & $>$1\,day & $>$1\,day & 58.75 & 58.46 & 58.85 \\
RE-GCN        & 44.86 & 39.82 & 46.75 & 65.69 & 59.98 & 68.70 \\
TANGO-TuckER  & 53.28 & 52.21 & 53.61 & 67.21 & 65.56 & 67.59 \\
TANGO-DistMult& 54.05 & 51.52 & 53.84 & 68.34 & 67.05 & 68.39 \\
CyGNet        & 45.50 & 50.48 & 50.79 & 63.47 & 64.26 & 65.71 \\
EvoKG         & 50.66 & 12.21 & 63.84 & 55.11 & 54.37 & 81.38 \\
HIP           & 54.71 & 53.82 & 54.73 & 67.55 & 66.32 & 68.49 \\
CENET         & 68.39 & 68.33 & 68.36 & \underline{84.13} & \underline{84.03} & 84.23 \\
TempValid     & \underline{83.19} & \underline{74.67} & \textbf{90.12} & 79.72 & 74.64 & \underline{84.78} \\
\midrule
\textbf{MS3M} & \textbf{84.25} & \textbf{81.20} & \underline{87.28}
              & \textbf{89.80} & \textbf{86.02} & \textbf{93.59} \\
\bottomrule
\end{tabular}
\end{table}

\subsection{Overall Performance Comparison (RQ1)}
\label{sec:overall_performance}

Table~\ref{tab:main_result} reports results on the event-centric suite. MS3M attains the highest Hits@1 on all four datasets and the highest MRR and Hits@3 on three of four, improving over the strongest baseline Hits@1 by $+3.6$ points on ICEWS14 ($0.355 \to 0.391$), $+2.2$ on ICEWS18 ($0.240 \to 0.262$), $+1.3$ on ICEWS05-15 ($0.408 \to 0.421$), and $+0.6$ on the million-edge GDELT stream ($0.153 \to 0.159$). Table~\ref{tab:wiki_yago_results} confirms strong generalization to coarse, long-horizon graphs: MS3M exceeds TempValid by $+6.53$ Hits@1 points on WIKI ($81.20\%$ vs.\ $74.67\%$) and leads every metric on YAGO ($+5.67$ MRR over CENET and $+8.81$ Hits@3 over TempValid).

\noindent\textbf{Dissecting Gains and the Hits@1 vs.\ Hits@10 Trade-off.} MS3M dominates top-rank precision (Hits@1 and MRR), whereas CognTKE records slightly higher Hits@10 on ICEWS18, ICEWS05-15, and GDELT. This reflects a clear distributional distinction: global--local smoothing in CognTKE produces flatter candidate posteriors favoring broad Top-10 recall, whereas MS3M's selective state-space gating and calibrated recurrence \emph{concentrate} mass onto the most causally supported tail. Cross-referencing Tables~\ref{tab:ablation} and~\ref{tab:evidence_type} attributes MS3M's gains to three complementary pillars: (i) calibrated recurrence supplies the largest precision lift on repeated $(s,r)$ contexts; (ii) multi-scale selective memory contributes $+0.014$ to $+0.021$ MRR by decoupling acute triggers from long-range regularities; and (iii) open-space generation with query-aware retrieval prevents probability collapse, driving $+25.8\%$ to $+32.1\%$ relative MRR gains on novel non-recurrent future entities across benchmarks.

\begin{table}[t]
\centering
\footnotesize
\setlength{\tabcolsep}{5pt}
\renewcommand{\arraystretch}{1.05}
\caption{Multi-seed stability (Mean $\pm$ Std across $3$ seeds) and one-tailed one-sample $t$-test significance of MS3M against the strongest reported baseline on each event-centric dataset.}
\label{tab:significance}
\resizebox{\textwidth}{!}{%
\begin{tabular}{@{}l|cc|cc|c@{}}
\toprule
\textbf{Dataset} & \textbf{Best Baseline MRR} & \textbf{MS3M MRR (Mean $\pm$ Std)} & \textbf{Best Baseline H@1} & \textbf{MS3M H@1 (Mean $\pm$ Std)} & \textbf{$p$-value} \\
\midrule
ICEWS14    & 0.458 (TempValid) & $0.4776 \pm 0.0019$ & 0.355 (TempValid) & $0.3904 \pm 0.0024$ & $0.0004$ \\
ICEWS18    & 0.342 (CognTKE)   & $0.3574 \pm 0.0011$ & 0.240 (CognTKE)   & $0.2615 \pm 0.0016$ & $0.0008$ \\
ICEWS05-15 & 0.519 (CognTKE)   & $0.5178 \pm 0.0014$ & 0.408 (CognTKE)   & $0.4216 \pm 0.0017$ & $0.0016$ (H@1) \\
GDELT      & 0.241 (CognTKE)   & $0.2433 \pm 0.0005$ & 0.153 (CognTKE)   & $0.1586 \pm 0.0008$ & $0.0074$ \\
\bottomrule
\end{tabular}%
}
\end{table}

\noindent\textbf{Statistical Significance.} Table~\ref{tab:significance} confirms tight multi-seed stability, with variance scaling inversely with test-set size (from $\pm 0.0019$ MRR on ICEWS14 down to $\pm 0.0005$ on GDELT). Even on GDELT, where the absolute MRR margin over CognTKE is $0.002$, the minimal variance across $305\text{K}$ test queries confirms significance ($p = 0.0074 < 0.01$), reinforced by a $+0.006$ Hits@1 gain.

\subsection{Ablation Study and Design Verification (RQ2)}
\label{sec:ablation}

\begin{table}[t]
\centering
\footnotesize
\setlength{\tabcolsep}{3pt}
\renewcommand{\arraystretch}{1.08}
\caption{Core component ablation study on four event-centric benchmarks (best in \textbf{bold}).}
\label{tab:ablation}
\resizebox{\textwidth}{!}{%
\begin{tabular}{@{}l|cccc|cccc|cccc|cccc@{}}
\toprule
\multirow{2}{*}{\textbf{Variant}}
& \multicolumn{4}{c}{\textbf{ICEWS14}}
& \multicolumn{4}{c}{\textbf{ICEWS18}}
& \multicolumn{4}{c}{\textbf{ICEWS05-15}}
& \multicolumn{4}{c}{\textbf{GDELT}} \\
\cmidrule(lr){2-5}\cmidrule(lr){6-9}\cmidrule(lr){10-13}\cmidrule(l){14-17}
& \textbf{MRR} & \textbf{H@1} & \textbf{H@3} & \textbf{H@10}
& \textbf{MRR} & \textbf{H@1} & \textbf{H@3} & \textbf{H@10}
& \textbf{MRR} & \textbf{H@1} & \textbf{H@3} & \textbf{H@10}
& \textbf{MRR} & \textbf{H@1} & \textbf{H@3} & \textbf{H@10} \\
\midrule
w/o Multi-Scale History           & 0.457 & 0.372 & 0.499 & 0.621 & 0.342 & 0.252 & 0.385 & 0.515 & 0.505 & 0.410 & 0.561 & 0.685 & 0.226 & 0.143 & 0.250 & 0.389 \\
w/o Selective State-Space Memory  & 0.464 & 0.379 & 0.502 & 0.625 & 0.348 & 0.255 & 0.394 & 0.526 & 0.509 & 0.414 & 0.563 & 0.689 & 0.228 & 0.145 & 0.251 & 0.390 \\
w/o Query-Aware Retrieval         & 0.471 & 0.385 & 0.513 & 0.635 & 0.350 & 0.257 & 0.393 & 0.528 & 0.505 & 0.406 & 0.559 & 0.690 & 0.237 & 0.154 & 0.261 & 0.400 \\
w/o Recurrence Calibration        & 0.429 & 0.344 & 0.466 & 0.598 & 0.336 & 0.244 & 0.376 & 0.509 & 0.482 & 0.391 & 0.525 & 0.655 & 0.203 & 0.136 & 0.224 & 0.342 \\
\midrule
\textbf{MS3M}                     & \textbf{0.478} & \textbf{0.391} & \textbf{0.520} & \textbf{0.642} & \textbf{0.358} & \textbf{0.262} & \textbf{0.403} & \textbf{0.539} & \textbf{0.517} & \textbf{0.421} & \textbf{0.570} & \textbf{0.698} & \textbf{0.243} & \textbf{0.159} & \textbf{0.266} & \textbf{0.407} \\
\bottomrule
\end{tabular}%
}
\end{table}

\noindent\textbf{Core Component Ablation.} We evaluate four controlled ablations (Table~\ref{tab:ablation}): (i) \textbf{w/o Multi-Scale History} retains only the finest stream ($K=1$); (ii) \textbf{w/o Selective State-Space Memory} replaces Equations~\eqref{eq:ssm_state}--\eqref{eq:ssm_out} with a parameter-matched non-selective exponential moving average plus a two-layer feed-forward projection; (iii) \textbf{w/o Query-Aware Retrieval} replaces masked attention ($\pi_i$) with uniform average pooling over $\bar{\mathbf{Z}}$; and (iv) \textbf{w/o Recurrence Calibration} removes $P_{\mathrm{rec}}$ and $\mathcal{L}_{\mathrm{rec}}$, scoring solely via $P_{\mathrm{gen}}$. Removing any pillar degrades every dataset: \emph{Recurrence Calibration} yields the largest drop ($-0.049$ MRR on ICEWS14, $-0.040$ on GDELT), followed by \emph{Multi-Scale History} ($-0.021$, $-0.017$), \emph{Selective State-Space Memory} ($-0.014$, $-0.015$), and \emph{Query-Aware Retrieval} ($-0.012$ on ICEWS05-15).

\begin{table}[t]
\centering
\footnotesize
\setlength{\tabcolsep}{5pt}
\renewcommand{\arraystretch}{1.05}
\caption{Verification of multi-scale construction, scale fusion, and memory recurrence choices on ICEWS14 and GDELT. Best results per block are in \textbf{bold}.}
\label{tab:design_choices}
\resizebox{\textwidth}{!}{%
\begin{tabular}{@{}ll|cc|cc@{}}
\toprule
\multirow{2}{*}{\textbf{Design Dimension}} & \multirow{2}{*}{\textbf{Strategy}} & \multicolumn{2}{c|}{\textbf{ICEWS14}} & \multicolumn{2}{c}{\textbf{GDELT}} \\
\cmidrule(lr){3-4}\cmidrule(l){5-6}
& & \textbf{MRR} & \textbf{H@1} & \textbf{MRR} & \textbf{H@1} \\
\midrule
\multirow{4}{*}{\textbf{Scale Construction}}
& Dilated Subsampling ($\rho=2$)            & 0.465 & 0.376 & 0.233 & 0.148 \\
& Strided 1D Convolution ($\rho=2$)         & 0.473 & 0.388 & 0.242 & 0.157 \\
& Exponential Time-Binning                  & 0.475 & 0.385 & 0.237 & 0.154 \\
& Mask-Aware Coarsening (Default)           & \textbf{0.478} & \textbf{0.391} & \textbf{0.243} & \textbf{0.159} \\
\midrule
\multirow{3}{*}{\textbf{Scale Fusion}}
& Learnable Static Weights ($\omega_k$)     & 0.476 & 0.389 & \textbf{0.244} & 0.158 \\
& Query-Adaptive Scale Attention            & \textbf{0.479} & 0.390 & \textbf{0.244} & \textbf{0.160} \\
& Uniform Equal Fusion (Default)            & 0.478 & \textbf{0.391} & 0.243 & 0.159 \\
\midrule
\multirow{3}{*}{\textbf{Memory Recurrence}}
& Standard GRU Encoder (Nonlinear Coupled)  & 0.466 & 0.380 & 0.235 & 0.153 \\
& Standard LSTM Encoder (Nonlinear Coupled) & 0.473 & 0.385 & 0.239 & 0.155 \\
& Diagonal Selective SSM (Default)          & \textbf{0.478} & \textbf{0.391} & \textbf{0.243} & \textbf{0.159} \\
\bottomrule
\end{tabular}%
}
\end{table}

\noindent\textbf{Architectural Design Verification.} Table~\ref{tab:design_choices} benchmarks MS3M's core choices against competitive alternatives on ICEWS14 and GDELT:
(1) \emph{Mask-aware coarsening} achieves the best accuracy across both regimes: dilated subsampling drops intermediate causal events ($-0.013$ MRR on ICEWS14, $-0.010$ on GDELT), exponential time-binning struggles under GDELT's uniform cadence ($0.237$), and strided 1D convolution matches GDELT closely ($0.242$) while slightly overfitting sparser ICEWS14 entities ($0.473$).
(2) \emph{Uniform equal fusion} stays within $0.001$ of learnable and query-adaptive fusion (edging query-adaptive attention on ICEWS14 Hits@1, $0.391$ vs.\ $0.390$) with zero added parameters, because per-scale exposure gates ($\mathbf{g}_i^{(k)}$), relation gating ($\boldsymbol{\eta}_r$), and masked attention ($\pi_i$) already execute fine-grained routing.
(3) \emph{Diagonal selective SSM} surpasses nonlinearly coupled GRU ($+0.012$ MRR on ICEWS14, $+0.008$ on GDELT) and LSTM ($+0.005$, $+0.004$), validating that input-dependent diagonal linear recurrences improve both predictive accuracy and optimization stability.

\subsection{Sensitivity Analysis (RQ3)}
\label{sec:sensitivity_efficiency}

\begin{figure}[t]
    \centering
    \includegraphics[width=\textwidth]{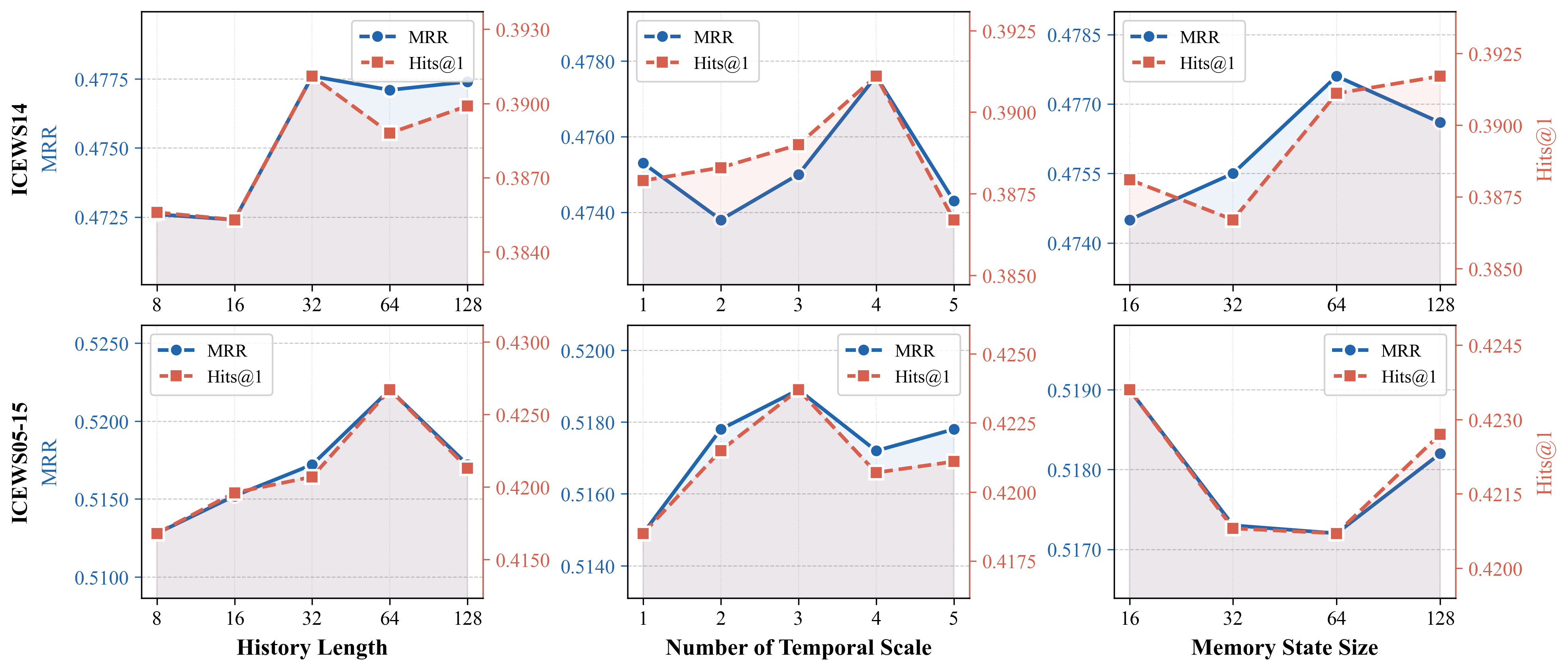}
    \caption{Hyperparameter sensitivity of MS3M on ICEWS14 and ICEWS05-15 across history length $L$, number of temporal scales $K$, and memory state size.}
    \label{fig:hyper_sensitivity}
\end{figure}

Figure~\ref{fig:hyper_sensitivity} evaluates MS3M across history length $L$, number of scales $K$, and memory state size on ICEWS14 and ICEWS05-15, showing smooth variation within a narrow band:
(1) \textbf{History Length $L$:} Short windows ($L \in \{8, 16\}$) truncate long-range regularities; accuracy peaks at $L=32$ on ICEWS14 and $L=64$ on longer-spanning ICEWS05-15 before plateauing at $L=128$ as distant noise accumulates.
(2) \textbf{Temporal Scales $K$:} Expanding from single-scale ($K=1$) to multi-scale memory steadily improves accuracy, peaking at $K=4$ on ICEWS14 and $K=3$ on ICEWS05-15; $K=5$ slightly over-compresses acute triggers.
(3) \textbf{Memory State Size:} Accuracy does not increase monotonically with capacity (peaking at $64$ on ICEWS14; strong at $16$ and rebounding at $128$ on ICEWS05-15), confirming that MS3M's gains stem from selective gating rather than brute-force capacity.

\subsection{Evidence-Stratified Analysis and Case Study (RQ4)}
\label{sec:temporal_evidence}

To verify whether MS3M genuinely reconciles heterogeneous temporal cues rather than inflating aggregate scores, we partition test queries at evaluation time into three mutually exclusive regimes using causal history $\mathcal{H}_{s,t}$, the recurrent set $\mathcal{C}_{s,r,t}$, and the recent window $\mathcal{R}_{s,t}$ (the $K_{\mathrm{short}}$ most recent events in $\mathcal{H}_{s,t}$): \textbf{Short-term Trigger} ($o^{*} \in \mathcal{R}_{s,t}$), \textbf{Long-term Recurrent} ($o^{*} \notin \mathcal{R}_{s,t},\ o^{*} \in \mathcal{C}_{s,r,t}$), and \textbf{Novel Entity} ($o^{*} \notin \mathcal{R}_{s,t},\ o^{*} \notin \mathcal{C}_{s,r,t}$).

\begin{table}[t]
\centering
\footnotesize
\setlength{\tabcolsep}{5pt}
\renewcommand{\arraystretch}{1.08}
\caption{Evidence-stratified evaluation (MRR) across short-term trigger, long-term recurrent, and novel-entity test queries (stratified strictly at evaluation time).}
\label{tab:evidence_type}
\begin{tabular}{@{}lc|cccc@{}}
\toprule
\textbf{Evidence Type} & \textbf{Ratio} & \textbf{DiffuTKG} & \textbf{RE-GCN} & \textbf{CognTKE} & \textbf{MS3M} \\
\midrule
\rowcolor{LightPurple}
\multicolumn{6}{@{}l}{\textbf{ICEWS14}} \\
Short-term Trigger  & 49\% & \textbf{0.778} & 0.687 & 0.712 & 0.758 \\
Long-term Recurrent & 13\% & 0.332 & 0.353 & 0.380 & \textbf{0.382} \\
Novel Entity        & 38\% & 0.109 & 0.093 & 0.083 & \textbf{0.144} \\
\midrule
\rowcolor{LightPurple}
\multicolumn{6}{@{}l}{\textbf{ICEWS18}} \\
Short-term Trigger  & 34\% & 0.627 & 0.586 & 0.615 & \textbf{0.657} \\
Long-term Recurrent & 23\% & 0.297 & 0.345 & \textbf{0.369} & 0.360 \\
Novel Entity        & 43\% & 0.091 & 0.093 & 0.070 & \textbf{0.117} \\
\bottomrule
\end{tabular}
\end{table}

Table~\ref{tab:evidence_type} reveals a decisive structural advantage. MS3M achieves the strongest overall balance across all three regimes---ranking first or close second on both short-term triggers ($0.657$ on ICEWS18, $0.758$ on ICEWS14) and long-term recurrences ($0.382$ on ICEWS14, $0.360$ on ICEWS18). Most notably, on \textbf{Novel Entity} queries---accounting for $38\%$ of ICEWS14 and $43\%$ of ICEWS18---MS3M outperforms every baseline by a wide margin ($0.144$ vs.\ $0.109$, $+32.1\%$ relative on ICEWS14; $0.117$ vs.\ $0.093$, $+25.8\%$ relative on ICEWS18), confirming that soft recurrence calibration exploits repetition without sacrificing novel-entity extrapolation.

\begin{table}[t]
\centering
\scriptsize
\setlength{\tabcolsep}{4pt}
\renewcommand{\arraystretch}{1.08}
\caption{Qualitative case study on ICEWS14 comparing open-space generation rank (Gen.) against recurrence-calibrated rank (Final) and the learned adaptive gate $\lambda$.}
\label{tab:case_study}
\begin{tabularx}{\textwidth}{>{\raggedright\arraybackslash\bfseries}p{0.13\textwidth} >{\raggedright\arraybackslash}X >{\raggedright\arraybackslash}X}
\toprule
\textbf{Case Type} & \textbf{Query and Historical Evidence} & \textbf{Prediction Behavior and Takeaway} \\
\midrule
Recent recurrent trigger
&
\textbf{Query:} $(\textit{China}, \textit{Express intent to cooperate}, ?, t)$ \newline
\textbf{Gold:} \textit{Japan} \newline
\textbf{Evidence:} Seen $26\times$ under $(s,r)$; last observed $1$ day prior.
&
\textbf{Rank (Gen.$\rightarrow$Final):} $37 \rightarrow 1$ \quad \textbf{$\lambda$:} $0.733$ \newline
\textbf{Top-3:} \textit{Japan}, \textit{Vietnam}, \textit{South Korea}. \newline
\textbf{Takeaway:} Recurrence calibration promotes a recently repeated, relation-consistent tail directly to rank $1$. \\
\midrule
Long-term recurrence
&
\textbf{Query:} $(\textit{South Africa}, \textit{Arrest or charge}, ?, t)$ \newline
\textbf{Gold:} \textit{Men (South Africa)} \newline
\textbf{Evidence:} Seen $125\times$ under $(s,r)$, absent from recent events.
&
\textbf{Rank (Gen.$\rightarrow$Final):} $98 \rightarrow 1$ \quad \textbf{$\lambda$:} $0.724$ \newline
\textbf{Top-3:} \textit{Men (South Africa)}, \textit{Police (South Africa)}, \textit{Citizen (Nigeria)}. \newline
\textbf{Takeaway:} Long-range frequency compensates for weak local cues, correcting the generation-only rank. \\
\midrule
Non-recurrent entity
&
\textbf{Query:} $(\textit{Guinea}, \textit{Engage in diplomatic cooperation}, ?, t)$ \newline
\textbf{Gold:} \textit{France} \newline
\textbf{Evidence:} Never seen under $(s,r)$; frequent tails: \textit{Morocco}, \textit{Sierra Leone}, \textit{China}, \textit{Angola}.
&
\textbf{Rank (Gen.$\rightarrow$Final):} $1 \rightarrow 5$ \quad \textbf{$\lambda$:} $0.530$ \newline
\textbf{Top-5:} \textit{China}, \textit{Angola}, \textit{Morocco}, \textit{Sierra Leone}, \textit{France}. \newline
\textbf{Takeaway:} Open-space generation surfaces the novel tail; moderated $\lambda$ applies a soft prior rather than hard masking. \\
\bottomrule
\end{tabularx}
\end{table}

Table~\ref{tab:case_study} illustrates this interplay qualitatively on ICEWS14. On the recent-trigger and long-term recurrence queries, MS3M activates high gates ($\lambda = 0.733$ and $0.724$), lifting the ground-truth entity from generation ranks $37$ and $98$ directly to rank $1$. On the non-recurrent query (\textit{France}), MS3M moderates the gate to $\lambda = 0.530$: while frequent historical partners receive calibrated mass (shifting \textit{France} from rank $1$ to $5$), the novel tail remains firmly among the top predictions rather than being masked out.

\section{Conclusion}
\label{sec:conclusion}

We presented \textbf{MS3M}, a Multi-Scale Selective State-Space Memory framework for temporal knowledge graph extrapolation. By organizing causal entity histories into a mask-aware multi-resolution pyramid, consolidating selective memory via input-dependent diagonal state-space recurrences in strictly linear $O(dL)$ time, and coupling open-space generation with frequency- and recency-calibrated recurrence, MS3M resolves the coupled trilemma of resolution mismatch, passive aggregation, and the recurrence--novelty dilemma. Across six benchmarks, MS3M achieves the highest Hits@1 on every dataset, state-of-the-art MRR on five, and substantial gains on novel future entities.

\noindent\textbf{Limitations and Future Work.} MS3M currently operates solely on structured quadruples without external textual semantics and uses a fixed coarsening factor ($\rho=2$) across entities. Future work will explore entity-adaptive temporal coarsening, multimodal event conditioning, and hardware-fused parallel associative-scan kernels for web-scale streaming temporal knowledge graphs.

\bibliographystyle{splncs04}
\bibliography{reference}

\end{document}